\documentclass[preprints,article,accept,pdftex,moreauthors]{Definitions/mdpi} 

\usepackage{graphicx}%
\usepackage{multirow}%
\usepackage{amsmath,amssymb,amsfonts}%
\usepackage{amsthm}%
\usepackage{mathrsfs}%
\usepackage[title]{appendix}%
\usepackage{xcolor}%
\usepackage{textcomp}%
\usepackage{manyfoot}%
\usepackage{booktabs}%
\usepackage{algorithm}%
\usepackage{algorithmicx}%
\usepackage{algpseudocode}%
\usepackage{listings}%

\usepackage{lineno}
\usepackage{color} 
\firstpage{1} 
\pubvolume{1}
\issuenum{1}
\articlenumber{0}
\pubyear{2025}
\copyrightyear{2025}
\datereceived{ } 
\daterevised{ } 
\dateaccepted{ } 
\datepublished{ } 
\hreflink{https://doi.org/} 

\Title{Event Data Association via Robust Model Fitting for Event-based Visual Object Tracking}

\TitleCitation{Event Data Association via Robust Model Fitting for Event-based Visual Object Tracking}

\Author{Haosheng Chen $^{1}$*\orcidA{}, Yue Wu $^{1}$\orcidB{} and Yidong Peng $^{1}$\orcidC{}}

\AuthorNames{Haosheng Chen, Yue Wu and Yidong Peng}

\isAPAStyle{%
       \AuthorCitation{Chen, H., Wu, Y., \& Peng, Y.}
         }{%
        \isChicagoStyle{%
        \AuthorCitation{Chen, Haosheng, Yue Wu, and Yidong Peng.}
        }{
        \AuthorCitation{Chen, H.; Wu, Y.; Peng, Y.}
        }
}

\address{%
$^{1}$ \quad The Chongqing Key Laboratory of Image Cognition, Chongqing University of Posts and Telecommunications, Chongqing 400065, China; chenhs@cqupt.edu.cn, wuyue@cqupt.edu.cn and pengyd@cqupt.edu.cn}

\corres{Correspondence: chenhs@cqupt.edu.cn}

\abstract{Event-based approaches, which are based on bio-inspired asynchronous event cameras, have achieved promising performance on various computer vision tasks. However, the study of the fundamental event data association problem is still in its infancy. In this paper, we propose a novel Event Data Association approach (called EDA) to explicitly address the data association problem. The proposed EDA seeks for event trajectories that best fit the event data, in order to perform unifying data association. In EDA, we first asynchronously gather the event data, based on its information entropy. Then, we introduce a deterministic model hypothesis generation strategy, which effectively generates model hypotheses from the gathered events, to represent the corresponding event trajectories. After that, we present a two-stage weighting algorithm, which robustly weighs and selects true models from the generated model hypotheses, through multi-structural geometric model fitting. Meanwhile, we also propose an adaptive model selection strategy to automatically determine the number of the true models. Finally, we use the selected true models to associate the event data, without being affected by sensor noise and irrelevant structures. We evaluate the performance of the proposed EDA on the remote object tracking task. The experimental results show the effectiveness of EDA under challenging scenarios, such as high speed, motion blur, and high dynamic range conditions.}


\begin{document}

\section{Introduction}

As one of the most delicate neural systems, biological retinas can precisely and efficiently capture object motions \cite{olveczky2003segregation}. Inspired by the biological retina, asynchronous event cameras, such as DVS \cite{lichtsteiner2008128}, DAVIS \cite{brandli2014240} and ATIS \cite{posch2011qvga}, have been proposed to mimic it. These bio-inspired silicon retina vision sensors have a very high dynamic range (higher than 100 dB) and an ultra-low latency (less than 1 ms). In particular, each pixel of the event camera can independently and asynchronously emit binary (i.e., \emph{On} or \emph{Off}) retinal events in responding to pixel intensity changes in the current environment. Pixel intensity changes are usually triggered by object and camera motions, under relatively smooth external illumination variations. Thus, the retinal events contain clear motion information. Moreover, thanks to the high dynamic range and low latency, event cameras are not affected by those challenges that inherently affect conventional frame-based cameras, such as motion blur and pixel saturation. As a result, event cameras are well suited for motion-related computer vision tasks, such as remote visual tracking.

\begin{figure*}[!t]
	\centering
	\includegraphics[width=0.85\linewidth]{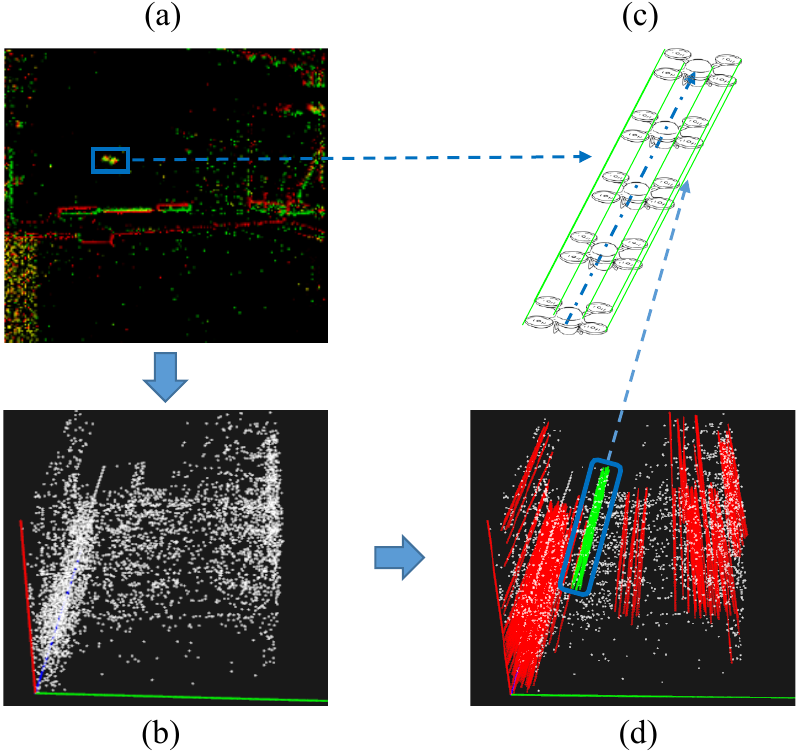}
	\caption{An illustration of event trajectories triggered by a flying drone. (a) The triggered retinal events on the image plane. The retinal events marked by the blue rectangle are triggered by the moving drone. (b) The triggered retinal events in the spatio-temporal domain. (c) The event trajectories triggered by the moving drone. (d) The event trajectories calculated by the proposed EDA in the 3D spatio-temporal domain. The green and red lines in (d) are triggered by the drone motion and the camera motion, respectively.}
	\label{fig:illustration}
\end{figure*}

Event-based methods have achieved promising performance on various tasks \cite{gallego2020event}. However, the study of the fundamental event data association problem is still challenging and in its infancy. Unlike a traditional camera, an event camera only sparsely emits binary (i.e., \emph{On} and \emph{Off}) retinal events at the edges within a scene, in responding to camera and object motions. Therefore, without the pixel intensity information, it is difficult to calculate event correspondences (i.e., event trajectories) in the temporal domain for event-based data association. Since data association is the key to various fundamental tasks (such as object tracking), the problem of event-based data association needs further investigation.

As the pioneering work on event-based data association, \cite{gallego2018unifying} proposes a contrast maximization framework to implicitly handle the data association problem based on the corresponding motion parameters. In that framework, the initial motion parameters for each event are empirically set to a range of predefined values for optimization. Since the predefined motion range may not cover all possible event motions, the final result may not be the optimal estimate.

However, \cite{gallego2018unifying} and other event-based data association methods \cite{zhu2017event, gallego2019focus} show that the events triggered by the same edge in the scene can be associated with each other using an event trajectory. Furthermore, they also show that the event trajectories, triggered at different edges of the scene, are nearly 3D lines over a short time interval in the spatio-temporal domain (see an example in Fig. \ref{fig:illustration}). Based on these observed facts, we explicitly formulate the event-based data association problem as a 3D event trajectory estimation problem in the spatio-temporal domain. Then, a novel event-based data association approach (called EDA) is proposed to perform explicit and accurate event-based data association. In the proposed EDA, the event trajectory estimation problem is effectively solved through the theory of robust multi-structural model fitting. To the best of our knowledge, it is the first time to introduce robust model fitting for event-based data association. The estimated multiple structures correspond to the event trajectories triggered by different motions, and they are used to associate the event data. To evaluate the effectiveness of the proposed EDA, we use EDA to deal with the problem of event-based data association in remote object tracking scenarios. Overall, this paper makes the following contributions:

\begin{itemize}
	\item We introduce a deterministic model hypothesis generation strategy to effectively generate model hypotheses in the spatio-temporal domain, spanned by retinal events. The generated model hypotheses are considered as event trajectory candidates for event data association.
	
	\item We propose a two-stage weighting algorithm, which robustly weighs the generated model hypotheses and selects true event trajectories from them through multi-structural model fitting, without being affected by sensor noise and irrelevant structures.
	
	\item We present an adaptive model selection strategy to automatically determine the number of the true event trajectories, based on the elbow value of the corresponding model hypothesis weights calculated by the proposed two-stage weighting algorithm.	
\end{itemize}

We extensively evaluate the proposed EDA on object tracking. The experimental results demonstrate the superiority of EDA over other state-of-the-art event-based tracking methods and several popular conventional tracking methods. In addition, the estimated true event trajectories corresponding to object motions are also provided for visualization.

\section{Related Work}
\label{sec:RelatedWork}
As mentioned previously, event cameras have shown their great potential on various computer vision tasks \cite{gallego2020event}. Different from conventional cameras that are synchronized to the global camera shutter, these bio-inspired vision sensors work in an asynchronous way with ultra low microsecond latencies in responding to object motions, which is inspired by the biological retina.

Thanks to the asynchronous nature, the high spatio-temporal resolution, and the superior HDR properties of event cameras, various event-based methods have been proposed for many computer vision tasks, especially for object tracking. In the field of visual remote tracking, conventional methods can be roughly categorized into three groups: UAV tracking methods\cite{xu2024sat,yang2024siamrhic,huang2024uav,jiang2023attention}, RGB tracking methods\cite{ye2024multimodal,liang2024joint,wang2023detector,hou2023siamese,Liang2020robust}, hyperspectral tracking methods\cite{gao2025feature,guo2024sptrack}, and RGB-T tracking methods\cite{xue2023siamcaf}. Compared with conventional methods, event-based tracking methods \cite{barranco2018real, camunas2017event, glover2017robust} show their superiority under fast motions and HDR scenes. The recently proposed event-based tracking methods have achieved state-of-the-art performance by leveraging motion compensation\cite{mitrokhin2018event}, asynchronous tracking-by-detection \cite{chen2019asynchronous}, and deep regression networks \cite{chen2020end}. Other notable motion-related event-based tasks include optical flow estimation \cite{Zhu_2019_CVPR}, camera motion estimation \cite{gallego2017event}, and motion segmentation \cite{stoffregen2019event}. In addition, there are also some works in relation to event-based feature extraction \cite{clady2015asynchronous}, classification \cite{lagorce2017hots}, and stereo vision \cite{stoffregen2019event}. From these event-based studies, we can see that the bio-inspired event camera can provide an effective and efficient way to solve various challenging vision problems, by exploiting the event data.

However, in the current event-based studies, most methods usually handle the fundamental event-based data association problem in implicit ways, which are designed for their specific tasks. As a result, event-based data association has not been effectively solved by the current event-based works. There are relatively few works focusing on this problem. As one of the pioneer works in this field, \cite{zhu2017event} proposes a probabilistic data association method based on the event-based features. Also, \cite{gallego2018unifying,gallego2019focus} present a unifying contrast maximization framework to implicitly perform data association. The retinal events are usually triggered by multiple motions, and they are greatly affected by a large amount of sensor noise. Due to these difficulties, the aforementioned methods use either indirect or implicit strategies to handle the event-based data association problem.  In this paper, we propose a unifying event data association approach (EDA), by which we explicitly handle the event-based data association problem through a robust multi-structural model fitting.

\begin{figure*}[!t]
	\centering
	\includegraphics[width=1.0\linewidth]{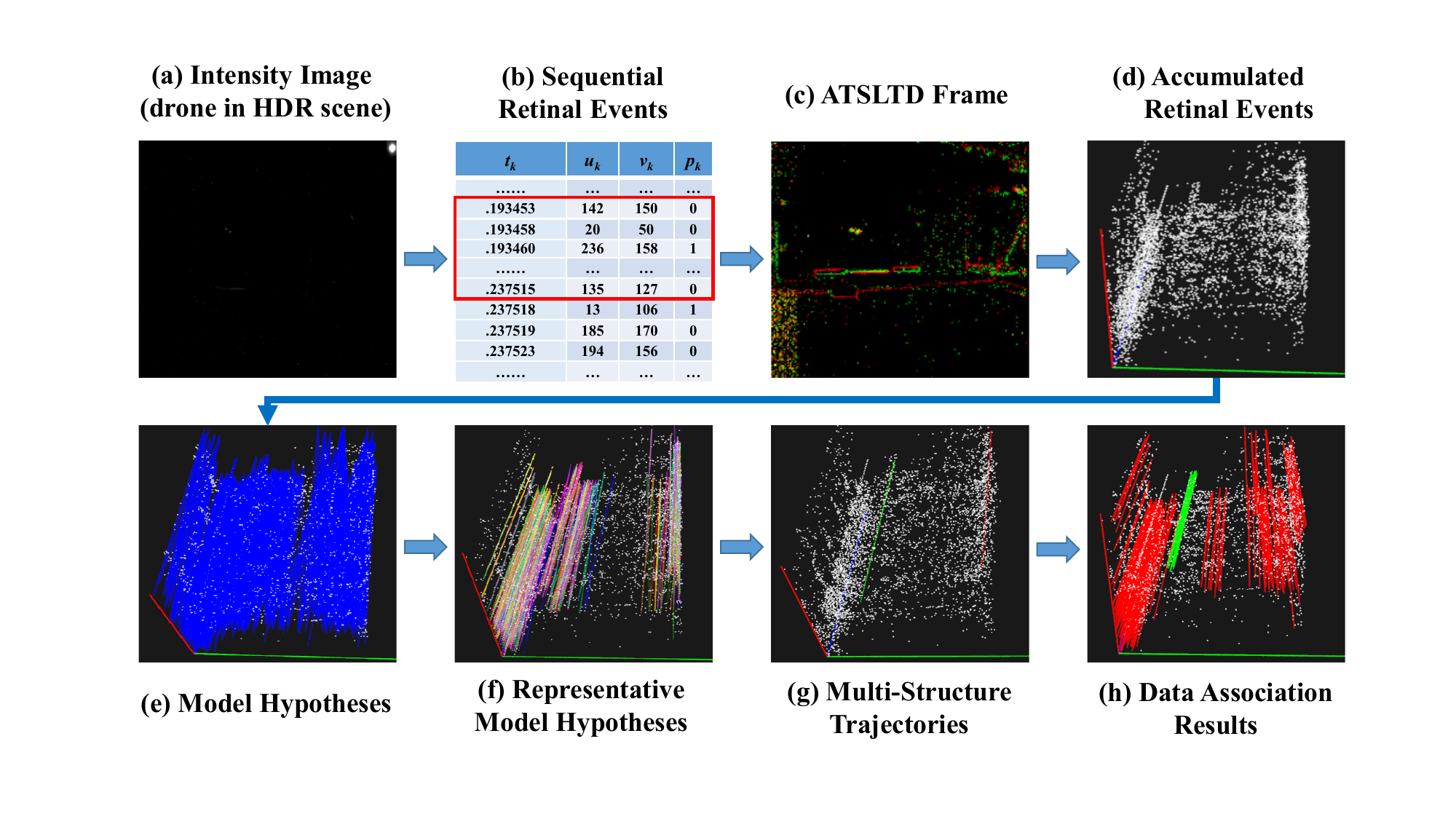}
	\caption{The pipeline of the proposed EDA. Note that the intensity image (a) is not used in EDA, and it is only for reference to show what conventional cameras would see in the low illumination scene. The three lines that are orthogonal to each other in (d)-(h) indicate the vertical (in red), horizontal (in green), and time (in blue) axes, respectively. The other lines in (e)-(h) are the event trajectory models that correspond to the their captions in (e)-(h).}
	\label{fig:pipeline}
\end{figure*}

Robust model fitting methods can accurately estimate the model instances (e.g., lines, homograph matrices, or fundamental matrices) in data contaminated with a large number of noise and outliers, and they have been applied in various areas. The most representative work in robust model fitting is the well-known RANSAC algorithm \cite{fischler1981random}. To improve the fitting performance, many robust fitting methods that use energy functions \cite{pham2014interacting, barath2018graph}, or use preference analysis \cite{magri2014t, magri2019fitting,xiao2020deterministic}, or use hypergraphs \cite{xiao2016hypergraph,wang2019searching}, or use geometric priors \cite{xiao2019superpixel,ma2019locality,ma2020image}, have been developed. However, there is no robust model fitting method proposed for event data. In this paper, we are the first to introduce robust model fitting to explicitly target the event association problem for object tracking.

\section{Proposed EDA Method}
\label{sec:Methods}

\subsection{Overview}
\label{subsec:MathodOverview}
In this section, we describe all the components of the proposed EDA approach. The pipeline of the proposed EDA is illustrated in Fig. \ref{fig:pipeline}. First, we describe the sequential retinal events, and introduce an asynchronous gathering phase to gather the sequential retinal events, as illustrated in Fig. \ref{fig:pipeline}(a)-(d). Then, we present the model hypothesis generation process, to effectively generate model hypotheses, as illustrated in Fig. \ref{fig:pipeline}(e)-(f). Finally, we give the details of the model hypothesis selection, to robustly perform the final data association, as illustrated in Fig. \ref{fig:pipeline}(g)-(h).

\subsection{Grouping Retinal Events}
\label{subsec:RetinalEvents}
During the recording of event data, each pixel of an event camera asynchronously and independently emits spatio-temporal retinal events, in responding to pixel intensity changes triggered by object motions. The $k$-th event $e_{k}$ of the emitted sequential retinal events $\mathcal{E}$ (see Fig. \ref{fig:pipeline}(b)) can be represented as follows:
\begin{equation}
	\label{eq1}
	e_{k} \doteq (t_{k}, u_{k}, v_{k}, p_{k}),
\end{equation}
where $t_{k}$ is the timestamp of the event, $u_{k}$ and $v_{k}$ are respectively the horizontal and vertical coordinates of the event, and $p_{k}$ indicates the polarity (i.e., \emph{On} or \emph{Off}). Increasing the pixel intensity will trigger an \emph{On} event, while decreasing the pixel intensity will trigger an \emph{Off} event. Since the changes of pixel intensity usually occur at the edges of objects in a scene due to object and camera motions, the corresponding retinal events are emitted accordingly, as shown in the ATSLTD frame of Fig. \ref{fig:pipeline}(c), where the green pixels represent the \emph{On} events and the red pixels represent the \emph{Off} events.

In this work, our goal is to calculate the correct event trajectories to associate the emitted retinal events. Since these event trajectories are approximately 3D lines over short time intervals (or small retinal event displacements), as pointed out by \cite{gallego2018unifying}, we need to determine the short time intervals over which we gather the sequential retinal events $\mathcal{E}$. Currently, event-based methods can be coarsely categorized into two types, i.e., event-by-event-based and groups-of-events-based, as discussed in \cite{gallego2020event}. A single event only contains very limited information, thus event-by-event-based methods usually rely on additional appearance information. In contrast, many event-based methods use groups of events as their inputs. However, these groups-of-events-based methods usually set the time interval to a constant value (or use a fixed event number as the threshold), which is not adaptive to different scenes. Moreover, gathering events over a constant time fails to leverage the asynchronous nature of the event data.

To keep the asynchronous nature of event data for benefiting high-level event-based tasks, we propose to leverage the asynchronous event-to-frame strategy in \cite{chen2019asynchronous} to asynchronously gather the sequential retinal events $\mathcal{E}$ into different sets for data association. According to that strategy, the sequential retinal events are accumulated to form an Adaptive Time-Surface with Linear Time Decay (ATSLTD) frame $\mathcal{F}$. Each of the incoming events will trigger an update at the corresponding pixel coordinates of the ATSLTD frame $\mathcal{F}$, thereby increasing the amount of information contained in the ATSLTD frame $\mathcal{F}$. Then, the amount of information in the ATSLTD frame is measured by the image entropy of $\mathcal{F}$. Here, the image entropy of $\mathcal{F}$ is quantitatively calculated by a Non-Zero Grid Entropy (NZGE) metric. After that, a set of image entropy values is collected from those ATSLTD frames that contain clear and sharp object contours. Here the clear and sharp object contours indicate relatively small event displacements, which keep the event trajectories straight. Finally, a confidence interval $[\alpha, \beta]$ of the collected image entropy values is calculated by using the Student's t-distribution. During the recording, the emitted sequential retinal events $\mathcal{E}$ will continuously update the ATSLTD frame $\mathcal{F}$. If the image entropy value of the ATSLTD frame $\mathcal{F}$ falls into the interval $[\alpha, \beta]$, the historical retinal events are accumulated for further processing. 

Suppose that the image entropy value of the accumulated retinal events collected from time $t_k$ falls into the calculated confidence interval $[\alpha, \beta]$ at time $t_k+T$. The corresponding accumulated retinal events are $\mathcal{E}_{t_k, t_k+T}$, and the next time interval starts from time $t_k+T$. By repeating this process, we can asynchronously gather the incoming retinal events into different sets for continuous event trajectory estimation and event-based data association.

\subsection{Generating Model Hypotheses}
\label{subsec:ModelHypothesisGeneration}
After the grouping process, we have obtained a set of accumulated retinal events $\mathcal{E}_{t_k, t_k+T}$, which is collected during the time interval $[t_k, t_k+T]$. The accumulated retinal events $\mathcal{E}_{k, k+T}$ can be represented in a 3D spatio-temporal space, as shown in Fig. \ref{fig:pipeline}(d). In Fig. \ref{fig:pipeline}(d), the orthogonal red, green, and blue lines are respectively the vertical, horizontal, and time axises, and each of the white voxels indicates a retinal event. In the 3D spatio-temporal space, each of the accumulated retinal events $\mathcal{E}_{k, k+T}$ is a 3D event voxel, regardless of its polarity. The $k$-th event $e_{k}=(t_{k}, u_{k}, v_{k}, p_{k})$ of $\mathcal{E}_{k, k+T}$ is represented by the $k$-th 3D event voxel ${\bf{e}}_{k} = (u_{k}, v_{k}, {t_k})$.

As mentioned previously, the expected event trajectories (i.e., the models), which best fit the event data, are 3D lines. Therefore, we propose an effective deterministic model hypothesis generation strategy to generate a set of 3D line hypotheses $\mathcal{L}_{t_k, t_k+T}$ from the accumulated retinal events $\mathcal{E}_{t_k, t_k+T}$ (some examples are shown in the blue lines in Fig. \ref{fig:pipeline}(e)). Each generated model hypothesis can be mathematically defined by its start event voxel and its end event voxel. For instance, the $k$-th hypothesis ${\bf l}_k$ is represented by its start event voxel ${\bf{e}}_{s}^{k} = (u_{s}^{k}, v_{s}^{k}, t_{s}^{k})$ and its end event voxel ${\bf{e}}_{e}^{k} = (u_{e}^{k}, v_{e}^{k}, t_{e}^{k})$ as follows:
\begin{equation}
	\label{eq2}
	{\bf l}_k = \left[ {\begin{array}{*{20}{c}}
			{{u_{s}^{k}} + \left( {{u_{e}^{k}} - {u_{s}^{k}}} \right)\lambda }\\
			{{v_{s}^{k}} + \left( {{v_{e}^{k}} - {v_{s}^{k}}} \right)\lambda }\\
			{{t_{s}^{k}} + \left( {{t_{e}^{k}} - {t_{s}^{k}}} \right)\lambda }
	\end{array}} \right],{\rm{    }}\forall \lambda  \in [0,1].
\end{equation}
During the model hypothesis generation process, we equally divide the accumulated retinal events $\mathcal{E}_{t_k, t_k+T}$ into $N_s$ time slices. Then, each of the generated model hypotheses is sampled using an event voxel pair, which includes an event voxel in the first time slice and an event voxel in the last time slice. As a result, different from traditional model hypothesis generation strategies (which usually randomly sample data to generate model hypotheses), the proposed model hypothesis generation strategy performs deterministic model hypothesis generation, based on guidance by the temporal information in the event data.

As illustrated in Fig. \ref{fig:illustration}, the event trajectories are usually triggered by the camera and object motions in the scene. Moreover, Fig. \ref{fig:illustration} also reveals the fact that the event trajectories, triggered by the same motion, are nearly parallel with each other, while the event trajectories triggered by different motions have relatively significant angle differences with each other. Therefore, we can refine the generated model hypotheses $\mathcal{L}_{t_k, t_k+T}$ to produce a few representative model hypotheses ${\hat {\mathcal L}_{{t_k},{t_k} + T}}$, as shown in Fig. \ref{fig:pipeline}(f), to relieve the computational burden. In Fig. \ref{fig:pipeline}(f), the representative model hypotheses triggered by different motions are marked in different colors. During the refining process, we calculate the cosine distances between two hypotheses in $\mathcal{L}_{t_k, t_k+T}$. The cosine distance between two hypotheses ${\bf l}_i$ and ${\bf l}_j$ is calculated as follows:
\begin{equation}
	\label{eq3}
	cosine\_dis ({{\bf{l}}_i},{{\bf{l}}_j}) = 1 - \frac{{{{\bf{l}}_i} \cdot {{\bf{l}}_j}}}{{\left\| {{{\bf{l}}_i}} \right\|\left\| {{{\bf{l}}_j}} \right\|}}.
\end{equation}
If the cosine distances among a set of hypotheses are close to zero, they are considered to be parallel with each other. Then, for each parallel hypothesis set, the model hypothesis that has the largest number of parallel hypotheses is selected as the representative hypothesis for the set, and the other hypotheses in the set are removed. After that, each representative model hypothesis is a unique event trajectory, and it is not parallel with the other representative hypotheses.

\subsection{Selecting Model Instances}
\label{subsec:ModelHypothesisSelection}
After the previous two steps, we have a set of representative model hypotheses ${\hat {\mathcal L}_{{t_k},{t_k} + T}}$. However, in most cases, since the event data is severely contaminated by sensor noise, the majority of the model hypotheses in ${\hat {\mathcal L}_{{t_k},{t_k} + T}}$ are bad model hypotheses. Here, we propose an effective two-stage weighting algorithm (called as TSW) to select the correct model hypotheses triggered by the multi-structural motions in the scene. Meanwhile, the TSW algorithm also removes the bad model hypotheses generated by the sensor noise.

During the first stage of the proposed TSW algorithm, we first calculate the residuals ${\mathcal R_{{t_k},{t_k} + T}}$ between the accumulated retinal events $\mathcal{E}_{t_k, t_k+T}$ and the representative model hypotheses ${\hat {\mathcal L}_{{t_k},{t_k} + T}}$ to select inliers $\mathcal{I}_{t_k, t_k+T}$ from $\mathcal{E}_{t_k, t_k+T}$. The residual ${\mathcal R_{{t_k},{t_k} + T}^{i,j}}$ between the $i$-th retinal event ${\bf e}_i$ in $\mathcal{E}_{t_k, t_k+T}$ and the $j$-th model hypothesis ${\bf l}_j$ in ${\hat {\mathcal L}_{{t_k},{t_k} + T}}$ is calculated by measuring the 3D voxel-to-line distance $d_{i,j}$ between ${\bf e}_i$ and ${\bf l}_j$, as follows:
\begin{equation}
	\label{eq4}
	{\mathcal R_{{t_k},{t_k} + T}^{i,j}} = {d_{i,j}} = \frac{{\left| {\left( {{{\bf{e}}_i} - {\bf{l}}_j^s} \right) \times \left( {{{\bf{e}}_i} - {\bf{l}}_j^e} \right)} \right|}}{{\left| {{\bf{l}}_j^e - {\bf{l}}_j^s} \right|}},
\end{equation}
where ${\bf{l}}_{j}^{s} = (u_{j}^{s}, v_{j}^{s}, t_{j}^{s})$ and ${\bf{l}}_{j}^{e} = (u_{j}^{e}, v_{j}^{e}, t_{j}^{e})$ are respectively the start event voxel of ${\bf l}_j$ and the end event voxel of ${\bf l}_j$. After the calculation, each representative model hypothesis has a vector of residuals, which is calculated with each retinal event in $\mathcal{E}_{t_k, t_k+T}$. To alleviate the problem of scale difference in the spatio-temporal space, the elements in the residual vector are normalized by the norm of the vector. 

Then, based on the residual ${\mathcal R_{{t_k},{t_k} + T}}$, for each of the model hypotheses in ${\hat {\mathcal L}_{{t_k},{t_k} + T}}$, the accumulated retinal events $\mathcal{E}_{t_k, t_k+T}$ can be divided into two sets: i.e., inliers $\mathcal{I}_{t_k, t_k+T}$ and outliers $\mathcal{O}_{t_k, t_k+T}$. For a reference model hypothesis in ${\hat {\mathcal L}_{{t_k},{t_k} + T}}$, if the residual (i.e., the 3D voxel-to-line distance) for a retinal event is less than an inlier noise scale $\tau$ (which can be estimated by using a robust scale estimator, such as IKOSE \cite{wang2011simultaneously}), the corresponding retinal event is considered as an inlier. Otherwise, the corresponding retinal event is considered as an outlier. If the inlier number of the reference model hypothesis is no more than two (i.e., only including the start voxel and the end voxel), this model hypothesis will be considered as noise and removed. After that, the reference model hypothesis will be weighted by measuring the dispersion of its inliers $\mathcal{I}_{t_k, t_k+T}$ on the time axis, which can be written as follows:
\begin{equation}
	\label{eq5}
	w = \frac{1}{{{N_{i}}}}\sum\limits_{i = 1}^{{N_{i}}} {{{({t_i} - {S_t}/2)}^2}},
\end{equation}
where $N_{i}$ is the number of inliers in $\mathcal{I}_{t_k, t_k+T}$, $t_i$ is the timestamp of the $i$-th inlier in $\mathcal{I}_{t_k, t_k+T}$, and ${S_t}$ is the length of the time axis. The first weighting stage in Eq. \ref{eq5} is based on the fact that the inliers of a true model hypothesis are supposed to be normally distributed along the time axis.

\begin{figure}[!t]
	\centering
	\includegraphics[width=0.9\linewidth]{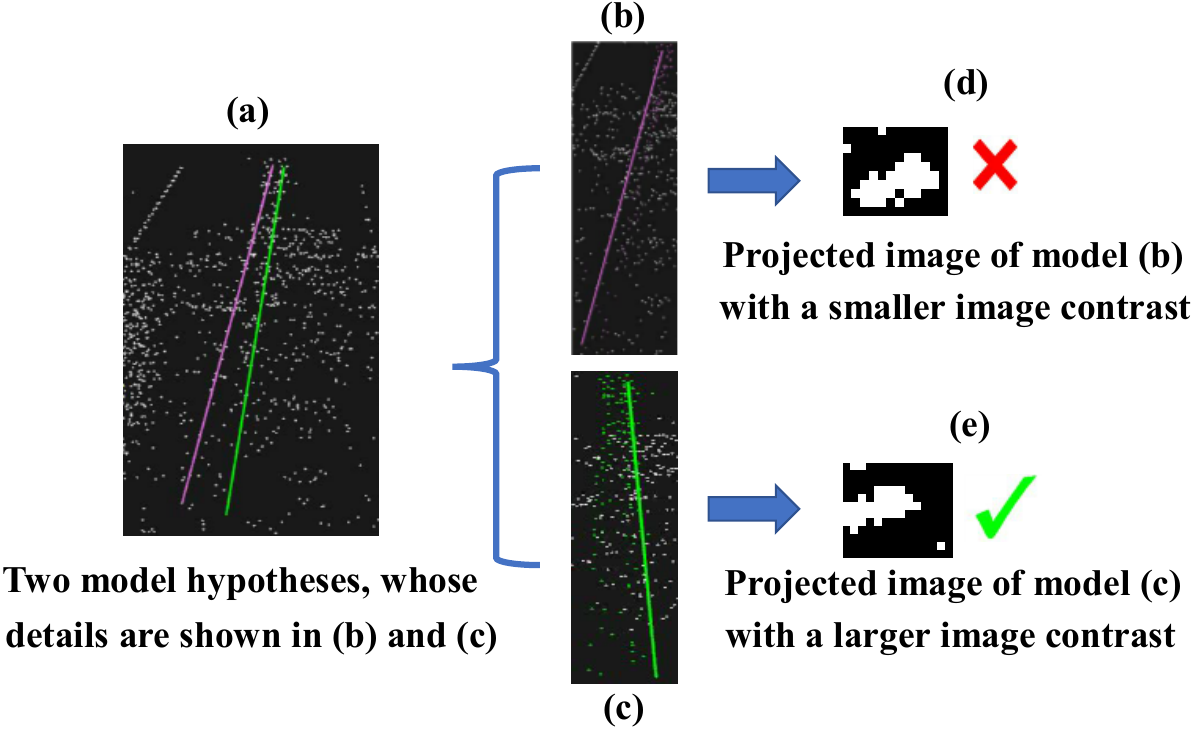}
	\caption{An illustration of the second stage of the proposed TSW algorithm. For the two model hypotheses in (a), compared with the projected image (d) from the sub-optimal model (b), the projected image (e) from the optimal model (c) has a larger image contrast. Thus, the model (c) is more likely to be a correct model hypothesis.}
	\label{fig:ImageContrast}
\end{figure}

After the first weighting stage, the second weighting stage is performed to reduce the influence of the pseudo-outliers (which are the inliers for a model hypothesis, but they are the outliers for the other model hypotheses). The second stage of the proposed TSW algorithm is based on the fact that the warped event image, projected by the true model hypothesis, has a larger image contrast compared with the false model hypotheses generated by the pseudo-outliers. An illustration of the second stage of the TSW algorithm is shown in Fig. \ref{fig:ImageContrast}. In the second stage, for the reference model hypothesis, we project its 3D spatio-temporal inliers in $\mathcal{I}_{t_k, t_k+T}$ to the reference 2D image plane according to the model hypothesis. After that, we crop the projected image to get a warped event image $\mathcal{P}_{t_k, t_k+T}$ using a bounding box, which is the minimum enclosing rectangle of the projected inliers in the 2D image plane. Moreover, we normalize the warped event image $\mathcal{P}_{t_k, t_k+T}$ and then calculate the image contrast of the normalized event image $\hat {\mathcal{P}}_{t_k, t_k+T}$ to adjust the weight $w$ of the reference model hypothesis, as follows:
\begin{equation}
	\label{eq6}
	\hat w = w*(1-\frac{1}{{{N_{\hat {\mathcal{P}}}}}}\sum\limits_{i,j} {{{({{\hat {\mathcal{P}}}_{t_k, t_k+T}^{i,j}} - {\mu _{\hat {\mathcal{P}}}})}^2}}),
\end{equation} 
where $N_{\hat {\mathcal{P}}}$ is the number of pixels of $\hat {\mathcal{P}}_{t_k, t_k+T}$, ${\hat {\mathcal{P}}}_{t_k, t_k+T}^{i,j}$ is the number of the projected inliers in the $i$-th row and the $j$-th column of $\hat {\mathcal{P}}_{t_k, t_k+T}$, and ${\mu _{\hat {\mathcal{P}}}}$ is the mean of $\hat {\mathcal{P}}_{t_k, t_k+T}$. After the two-stage weighting process, we have $N_m$ remaining model hypotheses together with their weights $\mathcal{\hat W} = \{ {\hat w_1},{\hat w_2}, \ldots ,{\hat w_{{N_m}}}\}$.

\begin{algorithm}[!t]
	\caption{The Proposed Event Data Association.}
	\label{alg1}  
	\begin{algorithmic}[1]  
		\Require The events $\mathcal{E}= \{ e_{1}, e_{2}, \cdots ,e_{n}\}$ collected from the start time $t_k$. 
		\State Initialize an ATSLTD frame $\mathcal{F}$ to a two-channel all-zero image with the resolution of the corresponding event camera; 		  
		\State Update $\mathcal{F}$ with the emitted events in $\mathcal{E}$ and accumulate the emitted events using the asynchronous event-to-frame strategy in Sec. \ref{subsec:RetinalEvents};
		\State Calculate the image entropy of $\mathcal{F}$ using the quantitative NZGE measure in the asynchronous event-to-frame strategy;
		\State Generate the confidence interval $[\alpha, \beta]$ for the calculated image entropy based on the Student's t-distribution or adopt pre-calculated intervals;
		\State Group the current accumulated events between $t_k$ and $t_k+T$ into a set of events $\mathcal{E}_{t_k, t_k+T}$ when the image entropy of $\mathcal{F}$ reach $[\alpha, \beta]$ at time $t_k+T$;
		\State Generate event trajectory model hypotheses $\mathcal{L}_{t_k, t_k+T}$ from the event set $\mathcal{E}_{t_k, t_k+T}$ using Eq. (\ref{eq2});
		\State Select the representative model hypotheses ${\hat {\mathcal L}_{{t_k},{t_k} + T}}$ from the generated event trajectory model hypotheses $\mathcal{L}_{t_k, t_k+T}$ using Eq. (\ref{eq3});
		\State Weigh the representative model hypotheses ${\hat {\mathcal L}_{{t_k},{t_k} + T}}$ based on the residuals ${\mathcal R_{{t_k},{t_k} + T}}$ between $\mathcal{E}_{t_k, t_k+T}$ and ${\hat {\mathcal L}_{{t_k},{t_k} + T}}$ using Eq. (\ref{eq4});
		\State Calculate an inlier noise scale $\tau$ for $\mathcal{E}_{t_k, t_k+T}$ using a robust scale estimator or a user-specified value;
		\State Select inliers $\mathcal{I}_{t_k, t_k+T}$ from $\mathcal{E}_{t_k, t_k+T}$ according to the inlier noise scale $\tau$;
		\State Weigh the representative model hypotheses ${\hat {\mathcal L}_{{t_k},{t_k} + T}}$ by measuring the dispersion of the inliers $\mathcal{I}_{t_k, t_k+T}$ on the time axis using Eq. (\ref{eq5});
		\State Weigh the representative model hypotheses ${\hat {\mathcal L}_{{t_k},{t_k} + T}}$ based on the weights in the previous step and the contrast of warped event images using Eq. (\ref{eq6});
		\State Estimate the number of true model hypotheses $N_s$ in the scene using the elbow weight of ${\hat {\mathcal L}_{{t_k},{t_k} + T}}$, and select true model hypotheses according to $N_s$;
		\Ensure The event trajectories associated by the true model instances.
	\end{algorithmic}
\end{algorithm}

The number of the true model instances $N_s$ in a scene (i.e., the number of the true motions) is usually unknown. Therefore, we propose an adaptive model selection strategy to automatically estimate it. The proposed strategy is based on the observation that the dispersion of weights between a true and a false model hypotheses is usually large, while the dispersion between a pair of the true model hypotheses is relatively small. Therefore, the elbow model weight that has the largest difference of weights among its neighbors, is critical for the estimation. In the proposed strategy, we sort all the remaining model hypotheses in an ascending order according to their weights $\mathcal{\hat W}$. Then, we calculate the difference of weights between every two adjacent sorted model hypotheses. The number of the true model structures $N_s$ is set to the first $k$, where the $k$-th difference value $d_{k}$ achieves the elbow value (i.e., the biggest difference value) among its four neighbors (i.e., $d_{k-2}$, $d_{k-1}$, $d_{k+1}$, and $d_{k+2}$).

Finally, the model hypotheses having the $N_s$ smallest weights are selected as the estimated model instances, as shown in Fig. \ref{fig:pipeline}(g), where the green and red lines are the estimated model instances. For each estimated model instance, we calculate its parallel model hypotheses in  $\mathcal{L}_{t_k, t_k+T}$, to perform the final data association. As a result, the inliers are associated with each other by their model structures, as shown in Fig. \ref{fig:pipeline}(h). Overall, the step-by-step description of the proposed EDA is given in Algorithm \ref{alg1}.

\section{Experiments}
\label{sec:Experiments}
As a fundamental task in computer vision, object tracking is still affected by high speed and high dynamic range problems. Therefore, in this work, we evaluate the performance of the proposed EDA on the task of object tracking.

\subsection{Parameter Settings}
For the parameters used in the proposed EDA, we adopt the same confidence interval $[\alpha, \beta]$ as that of ETD \cite{chen2019asynchronous} for fair competition. The number of the time slices $N_s$ is empirically set to 10 to achieve the trade-off between accuracy and efficiency. The inlier noise scale $\tau$ is fixed to 0.01 for all the experiments.

\subsection{Evaluation Protocol}
For the object tracking task, since the event data is associated between two adjacent frames, we employ an evaluation protocol of frame-wise tracking to evaluate the performance of the proposed EDA approach. For the frame-wise tracking, the evaluation of these competing methods is based on object pairs, each of which includes two object regions on two adjacent frames corresponding to the same object. During the evaluation, we treat each of the object pairs as a tracking instance in the corresponding frame.

The test object tracking sequences include the \emph{shapes\_translation}, \emph{shapes\_6dof}, \emph{poster\_6dof}, and \emph{slider\_depth} sequences from the popular Event Camera Dataset (ECD) \cite{mueggler2017event} and the \emph{fast\_drone}, \emph{light\_variations}, \emph{what\_is\_background}, and \emph{occlusions} sequences from the challenging Extreme Event Dataset (EED) \cite{mitrokhin2018event}. In the four sequences from the ECD dataset, the first three sequences have an increasing motion speed. The fourth sequence has a constant motion speed. For the four sequences, the corresponding object textures vary from simple Black\&White shapes to complicated artifacts. By using these four sequences, we are mainly concerned with the performance of all competing methods for a variety of object motions, especially for fast motion, and for different object shapes and textures. For the four sequences from the EED dataset, the first three sequences record a fast moving drone under low illumination environments. The fourth sequence records a moving ball with a net as the foreground. By using these four sequences, we evaluate the performance of the competing trackers in low illumination conditions and occlusion situations.

For the evaluation metrics, we follow the frame-wise tracking protocol in \cite{chen2020end} and use the Average Overlap Rate (AOR) and the Average Robustness (AR) to respectively measure the precision and the robustness of all the competing object tracking methods, as follows:
\begin{equation}
	\label{eq:aor}
	{AOR} = \frac{1}{{{N_{rep}}}}\frac{1}{{{N_{pair}}}}\sum\limits_{i = 1}^{{N_{rep}}} {\sum\limits_{j = 1}^{{N_{pair}}} {\frac{{O_{i,j}^E \cap O_{i,j}^G}}{{O_{i,j}^E \cup O_{i,j}^G}}}},
\end{equation}
\begin{equation}
	\label{eq:ar}
	{AR} = \frac{1}{{{N_{rep}}}}\frac{1}{{{N_{pair}}}}\sum\limits_{i = 1}^{{N_{rep}}} {\sum\limits_{j = 1}^{{N_{pair}}} {succes{s_{i,j}}}},
\end{equation}
where $O_{i,j}^{E}$ is the estimated bounding box in the $i$-th round of the evaluation for the $j$-th object pair. $O_{i,j}^{G}$ is the corresponding ground truth. ${N_{rep}}$ is the number of times that we repeat the evaluation. ${N_{pair}}$ is the number of object pairs in the current sequence. We set the value of ${N_{rep}}$ to 5 for all the following experiments. $succes{s_{i,j}}$ indicates that whether the tracking in the $i$-th round for the $j$-th pair is successful or not (0 means failure and 1 means success). If the AOR score obtained by a tracker for one object pair is under 0.5, we consider it as a tracking failure case.

\begin{table*}[!t]
	\renewcommand\tabcolsep{5.0pt}
	\caption{The AOR results obtained by the eight competitors and our EDA on the ECD dataset. The best results are highlighted in \textbf{bold}.}
	\small
	\begin{center}
		\begin{tabular}{|c|c|c|c|c|c|c|c|c|}
			\hline
			Method & \emph{shapes\_translation} & \emph{shapes\_6dof} & \emph{poster\_6dof} & \emph{slider\_depth} \\ \hline\hline
			SiamFC \cite{bertinetto2016fully}  & 0.812          & 0.835          & 0.830          & 0.909          \\ \hline
			ECO \cite{danelljan2017eco}        & 0.823          & 0.847          & 0.846          & 0.947          \\ \hline
			SiamRPN++ \cite{Li_2019_CVPR}      & 0.790          & 0.779          & 0.753          & 0.907          \\ \hline
			ATOM \cite{Danelljan_2019_CVPR}    & 0.815          & 0.803          & 0.835          & 0.897          \\ \hline
			ECO-E \cite{danelljan2017eco}      & 0.821          & 0.834          & 0.783          & 0.771          \\ \hline
			E-MS \cite{barranco2018real}       & 0.675          & 0.612          & 0.417          & 0.447          \\ \hline
			ETD \cite{chen2019asynchronous}    & 0.843          & 0.852          & 0.824          & 0.903          \\ \hline
			RMRNet \cite{chen2020end}          & 0.836          & \textbf{0.866} & 0.859          & 0.915          \\ \hline
			EDA(Ours)                          & \textbf{0.896} & \textbf{0.866} & \textbf{0.872} & \textbf{0.969} \\ \hline
		\end{tabular}
	\end{center}
	\label{tab:ecd_aor_results}
\end{table*}

\begin{table*}[!t]
	\renewcommand\tabcolsep{5.0pt}
	\caption{The AR results obtained by the eight competitors and our EDA on the ECD dataset. The best results are highlighted in \textbf{bold}.}
	\small
	\begin{center}
		\begin{tabular}{|c|c|c|c|c|c|c|c|c|}
			\hline
			Method & \emph{shapes\_translation} & \emph{shapes\_6dof} & \emph{poster\_6dof} & \emph{slider\_depth} \\ \hline\hline
			SiamFC \cite{bertinetto2016fully}  & 0.940          & 0.968          & 0.956          & \textbf{1.000} \\ \hline
			ECO \cite{danelljan2017eco}        & 0.943          & 0.969          & 0.960          & \textbf{1.000} \\ \hline
			SiamRPN++ \cite{Li_2019_CVPR}      & 0.942          & 0.972          & 0.899          & \textbf{1.000} \\ \hline
			ATOM \cite{Danelljan_2019_CVPR}    & 0.945          & 0.974          & 0.961          & \textbf{1.000} \\ \hline
			ECO-E \cite{danelljan2017eco}      & 0.941          & 0.960          & 0.878          & 0.993          \\ \hline
			E-MS \cite{barranco2018real}       & 0.768          & 0.668          & 0.373          & 0.350          \\ \hline
			ETD \cite{chen2019asynchronous}    & \textbf{0.998} & \textbf{0.998} & \textbf{0.995} & \textbf{1.000} \\ \hline
			RMRNet \cite{chen2020end}          & 0.951          & 0.980          & 0.962          & \textbf{1.000} \\ \hline
			EDA(Ours)                          & \textbf{0.998} & \textbf{0.998} & 0.983          & \textbf{1.000} \\ \hline
		\end{tabular}
	\end{center}
	\label{tab:ecd_ar_results}
\end{table*}

\begin{table*}[!t]
	\renewcommand\tabcolsep{5.0pt}
	\caption{The AOR results obtained by the eight competitors and our EDA on the EED dataset. The best results are highlighted in \textbf{bold}.}
	\small
	\begin{center}
		\begin{tabular}{|c|c|c|c|c|c|c|c|c|}
			\hline
			Method & \emph{fast\_drone} & \emph{light\_variations} & \emph{what\_is\_background} & \emph{occlusions} \\ \hline\hline
			SiamFC \cite{bertinetto2016fully}  & 0.766          & 0.772          & 0.712          & 0.148          \\ \hline
			ECO \cite{danelljan2017eco}        & 0.830          & 0.782          & 0.675          & 0.209          \\ \hline
			SiamRPN++ \cite{Li_2019_CVPR}      & 0.717          & 0.497          & 0.653          & 0.096          \\ \hline
			ATOM \cite{Danelljan_2019_CVPR}    & 0.763          & 0.652          & 0.725          & 0.387          \\ \hline
			ECO-E \cite{danelljan2017eco}      & 0.728          & 0.685          & 0.099          & 0.308          \\ \hline
			E-MS \cite{barranco2018real}       & 0.313          & 0.325          & 0.362          & 0.356          \\ \hline
			ETD \cite{chen2019asynchronous}    & 0.817          & 0.859          & 0.653          & 0.608          \\ \hline
			RMRNet \cite{chen2020end}          & 0.892          & 0.802          & 0.202          & \textbf{0.716} \\ \hline
			EDA(Ours)                          & \textbf{0.913} & \textbf{0.894} & \textbf{0.752} & 0.702          \\ \hline
		\end{tabular}
	\end{center}
	\label{tab:eed_aor_results}
\end{table*}

\begin{table*}[!t]
	\renewcommand\tabcolsep{5.0pt}
	\caption{The AR results obtained by the eight competitors and our EDA on the EED dataset. The best results are highlighted in \textbf{bold}.}
	\small
	\begin{center}
		\begin{tabular}{|c|c|c|c|c|c|c|c|c|}
			\hline
			Method & \emph{fast\_drone} & \emph{light\_variations} & \emph{what\_is\_background} & \emph{occlusions} \\ \hline\hline
			SiamFC \cite{bertinetto2016fully}  & \textbf{1.000} & 0.947          & 0.833          & 0.000          \\ \hline
			ECO \cite{danelljan2017eco}        & \textbf{1.000} & 0.934          & 0.750          & 0.333          \\ \hline
			SiamRPN++ \cite{Li_2019_CVPR}      & 0.941          & 0.500          & 0.833          & 0.167          \\ \hline
			ATOM \cite{Danelljan_2019_CVPR}    & \textbf{1.000} & 0.921          & \textbf{0.917} & 0.500          \\ \hline
			ECO-E \cite{danelljan2017eco}      & 0.882          & 0.803          & 0.000          & 0.333          \\ \hline
			E-MS \cite{barranco2018real}       & 0.307          & 0.321          & 0.360          & 0.353          \\ \hline
			ETD \cite{chen2019asynchronous}    & 0.941          & 0.934          & 0.807          & 0.647          \\ \hline
			RMRNet \cite{chen2020end}          & \textbf{1.000} & 0.947          & 0.083          & \textbf{0.833} \\ \hline
			EDA(Ours)                          & \textbf{1.000} & \textbf{0.966} & \textbf{0.917} & \textbf{0.833} \\ \hline
		\end{tabular}
	\end{center}
	\label{tab:eed_ar_results}
\end{table*}

\subsection{Event-Based Object Tracking}

\begin{figure*}[!t]
	\centering
	\includegraphics[width=1.0\linewidth]{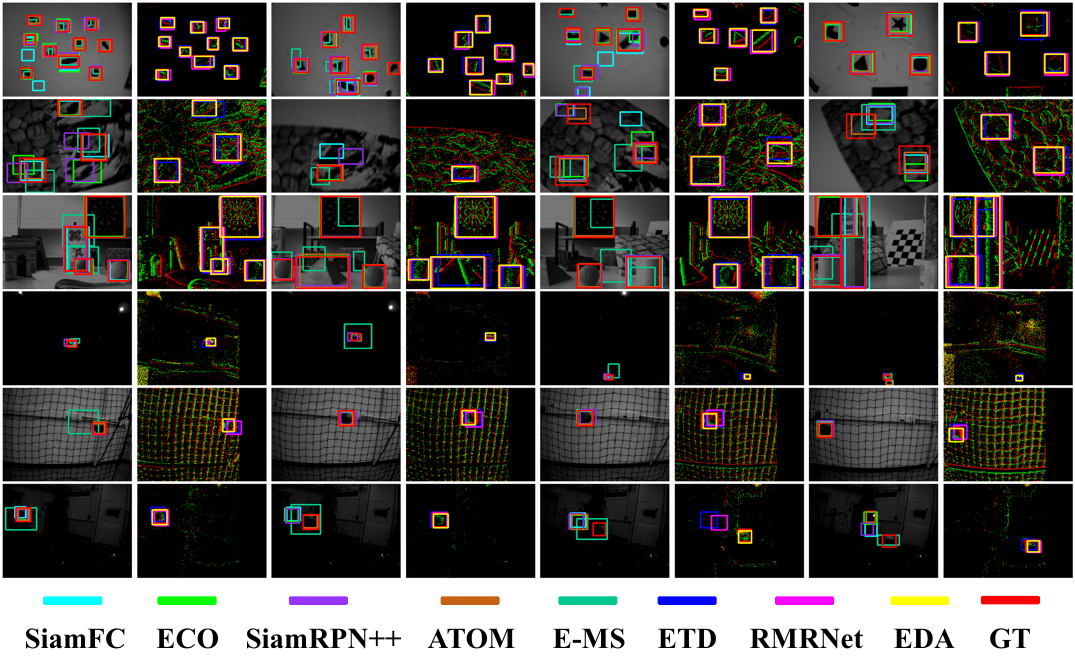}
	\caption{Quantitative tracking results obtained by SiamFC, ECO, SiamRPN++, ATOM, E-MS, ETD, RMRNet, and the proposed EDA. Each row represents a sequence. From top to bottom, the corresponding sequences are \emph{shape\_6dof}, \emph{poster\_6dof}, \emph{slider\_depth}, \emph{light\_variations}, \emph{what\_is\_background}, and \emph{occlusions}, respectively. From left to right, the 2nd, 4th, 6th, and 8th columns show the results obtained by ETD, RMRNet, and EDA with the ground truth (abbreviated as GT). The 1st, 3rd, 5th, and 7th columns show the results obtained by the other competing methods with the GT. Best viewed in color.}
	\label{fig:results}
\end{figure*}

In this work, we compare the proposed EDA with eight popular tracking methods, including SiamFC \cite{bertinetto2016fully}, ECO \cite{danelljan2017eco}, SiamRPN++ \cite{Li_2019_CVPR}, ATOM \cite{Danelljan_2019_CVPR}, E-MS \cite{barranco2018real}, ETD \cite{chen2019asynchronous}, RMRNet \cite{chen2020end}, and an event-based variants of ECO (called as ECO-E). Among these trackers, SiamFC, ECO, SiamRPN++, and ATOM are the popular conventional tracking methods. E-MS, ETD, and RMRNet are the state-of-the-art event-based tracking methods. ECO-E is an event-based variant of ECO, which uses TSLTD frames as its inputs. Thus, ECO-E is adopted to evaluate the performance of ECO on the event data. Moreover, E-MS is extended to support the bounding box-based object tracking.

The quantitative results are given in Table \ref{tab:ecd_aor_results}, Table \ref{tab:ecd_ar_results}, Table \ref{tab:eed_aor_results}, and Table \ref{tab:eed_ar_results}. We also provide some qualitative results obtained by SiamFC, ECO, SiamRPN++, ATOM, E-MS, ETD, RMRNet, and our EDA in Fig. \ref{fig:results}. From the tables and the figure, we can see that our EDA achieves the best performance on most of the sequences except for the \emph{poster\_6dof} and \emph{occlusions} sequences, on which our EDA has obtained the second best AR and AOR, respectively. This is because that, these two sequences (i.e., \emph{poster\_6dof} and \emph{occlusions}) include considerable object pairs that are almost out-of-view or highly occluded. These object pairs slightly affect the robustness of EDA compared with the edge-based ETD tracker and the motion estimation-based RMRNet tracker. The superior performance achieved by EDA is largely attributed to the proposed two-stage weighting algorithm, which accurately selects true event trajectories without being affected by the sensor noise.

\begin{figure*}[!t]
	\centering
	\includegraphics[width=1.0\linewidth]{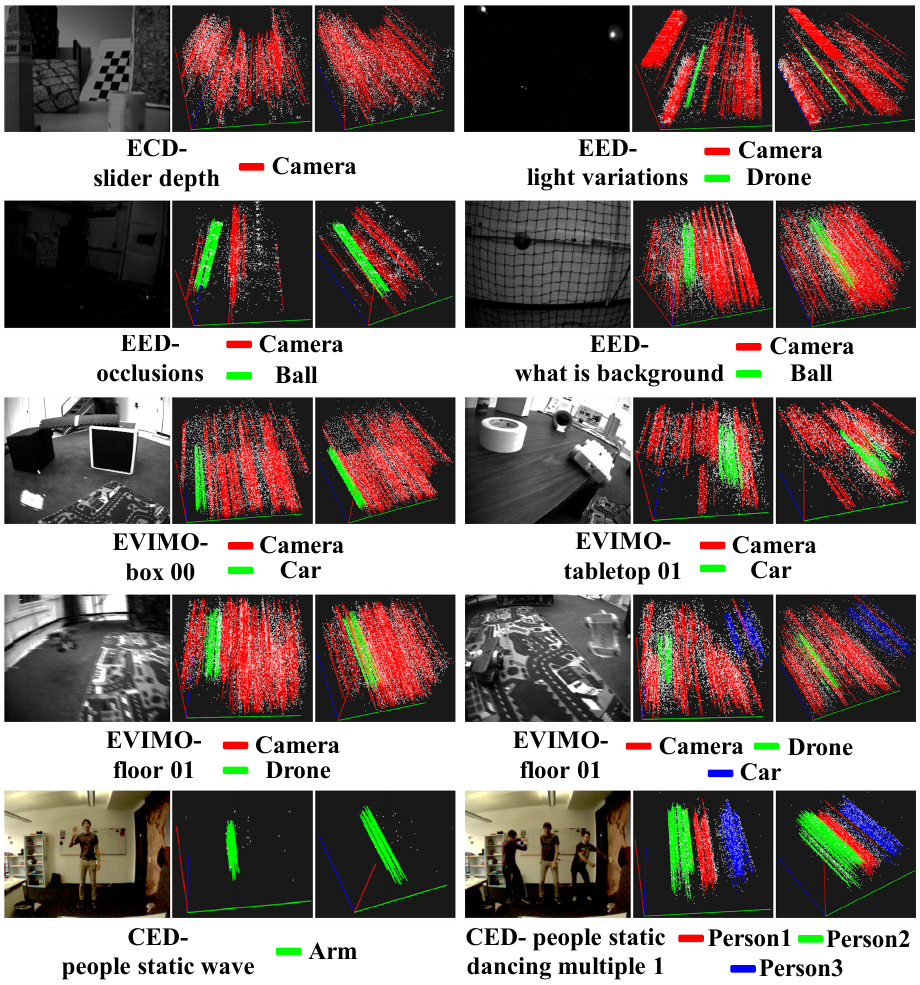}
	\caption{Event trajectory estimation results obtained by the proposed EDA approach. The event trajectories marked by different colors are triggered by different motions. The names and the corresponding motions of the video sequences are listed in the bottom of the corresponding event trajectory results, respectively. For each sequence, we show the intensity image of the scene (only for reference, not used in the proposed EDA) and two views of its event trajectory results. Best viewed in color.}
	\label{fig:segmotion_results}
\end{figure*}

As for the competitors, SiamFC, ECO, SiamRPN++, and ATOM also achieve competitive results on some sequences of the ECD dataset. However, they show inferior tracking performance on the sequences with fast motion and low illumination conditions (such as the \emph{light\_variations} and \emph{occlusions} sequences). They usually lose the tracked objects, due to the influence of motion blur and HDR conditions. In contrast, the three event-based competitors (i.e., ECO-E, E-MS, and ETD) can effectively handle fast motions and HDR scenes, by leveraging the good properties of the event data. Nevertheless, E-MS is less effective in handling the challenges of complicated object textures and cluttered backgrounds (e.g., on the \emph{poster\_6dof} and \emph{slider\_depth} sequences), and it is sensitive to sensor noise. Compared with E-MS, the ETD and RMRNet trackers achieve very competitive results on most of the sequences. However, due to the unstable performance during the processes of the edge-based object proposal generation and motion estimation, ETD and RMRNet achieve inferior precision performance compared with the proposed EDA. Moreover, compared with ECO, ECO-E has achieved much inferior performance, which indicates that directly applying conventional tracking methods to event data is less effective.

We also provide the event trajectory results obtained by the proposed EDA on some representative sequences from the ECD and EED datasets. Furthermore, in order to evaluate the performance of EDA under more tracking scenarios with challenges, we test it on some representative sequences from the Color Event Dataset (CED) \cite{scheerlinck2019ced} and the recently released EVIMO dataset \cite{mitrokhin2019ev}. Here, it should be noted that there is no ground truth that is publicly available for the CED and EVIMO datasets for the task of object tracking. Therefore, we only provide the event trajectory results obtained by EDA for qualitative evaluation. The qualitative results are shown in Fig. \ref{fig:segmotion_results}. 

From the results in Fig. \ref{fig:segmotion_results}, we can see that the proposed EDA has accurately estimated most of the true event trajectories triggered by the camera and object motions through robust model fitting. In particular, it is worth noting that, with the help of the proposed adaptive model selection strategy, the numbers of the true models for all the video sequences in Fig. \ref{fig:segmotion_results} are correctly estimated by the proposed EDA.  Moreover, EDA works well when dealing with various challenging situations including HDR scenes (i.e., the EED-light\_variations and EED-occlusions sequences), fast motions (i.e., all the three sequences from the EED dataset and the EVIMO-floor\_01 sequence), and multi-structural motions (i.e., the EVIMO-floor\_01 and CED-people\_static\_dancing\_multiple\_1 sequences).

\section{Discussion}

\begin{figure}[!t]
		\begin{adjustwidth}{-\extralength}{0cm}
			\centering
		\begin{flushright}
			\subfloat[\centering]{\includegraphics[width=7.0cm]{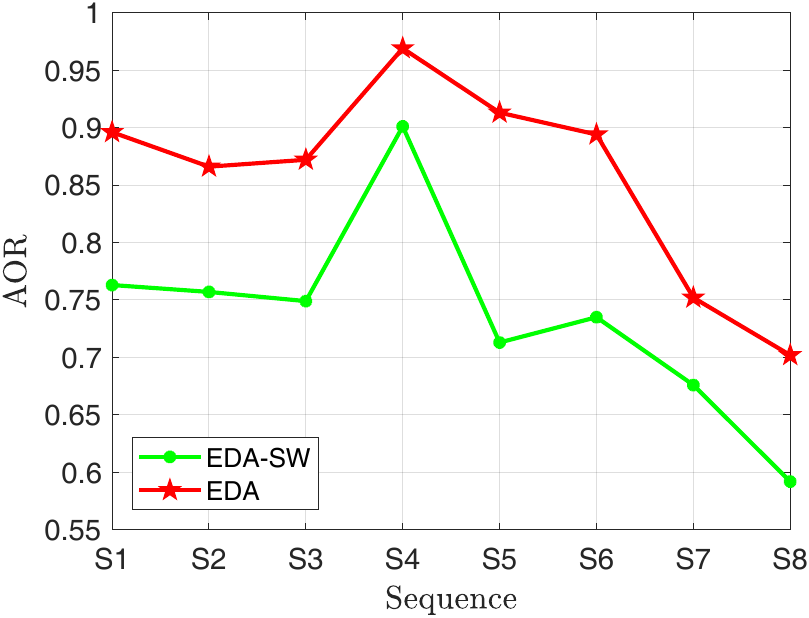}}
			\subfloat[\centering]{\includegraphics[width=7.0cm]{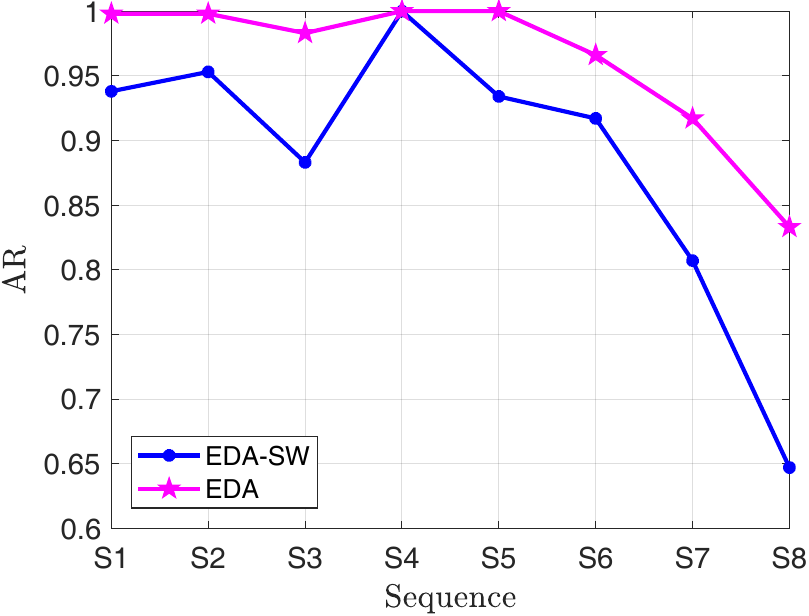}}
	    \end{flushright}
		\end{adjustwidth}
	\caption{The AOR and AR results obtained by the proposed EDA and its variant EDA-SW on the eight sequences in the ECD and EED datasets. The corresponding eight sequences S1 to S8 are the sequences from \emph{shapes\_translation} to \emph{occlusions}.}
	\label{fig:ablation}
\end{figure}

\begin{figure}[!t]
		\begin{adjustwidth}{-\extralength}{0cm}
			\centering
		\begin{flushright}
			\subfloat[\centering]{\includegraphics[width=7.0cm]{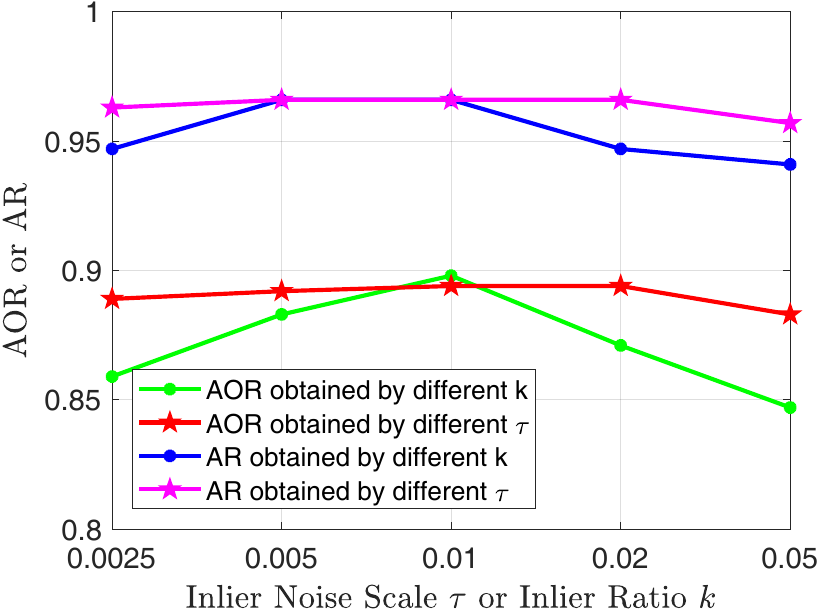}}
			\subfloat[\centering]{\includegraphics[width=7.0cm]{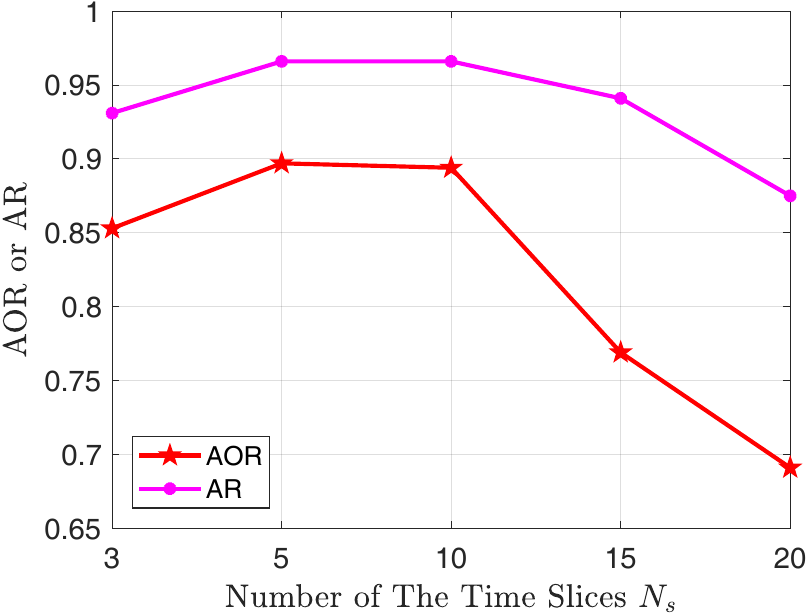}}
		\end{flushright}
		\end{adjustwidth}
	\caption{Parameter analysis on the inlier noise scale $\tau$, the inlier ratio $k$ in IKOSE, and the number of the time slices $N_s$. The AOR and AR results obtained by the proposed EDA with different values of the inlier noise scale $\tau$ and different values of the inlier ratio $k$ are shown in the left image. The AOR and AR results obtained by the proposed EDA with different numbers of the time slices $N_s$ are shown in the right image.}
	\label{fig:parameter}
\end{figure}

\subsection{Ablation Study}
\label{subsec:ablationstudy}
To evaluate the proposed EDA on visual tracking, we need to calculate the corresponding object bounding box based on the event trajectories associated by EDA. According to the frame-wise tracking protocol, for each tracking instance, we have the ground truth bounding box of the tracked object at the current frame. EDA needs to estimate the corresponding object bounding box of the tracked object at the next frame. To achieve this goal, we project those retinal events within the current bounding box to the reference image plane at the timestamp of the next frame using the estimated event trajectories calculated by EDA. Then, we use the minimum enclosing rectangle of the projected events in the reference plane as the estimated bounding box at the current frame.

It is worth pointing out that the proposed EDA is the first that introduces robust model fitting to solve the event data association problem for object tracking. Since EDA does not rely on any existing event data association methods or object tracking methods, there is no baseline method for it. In EDA, only the second stage of the TSW algorithm can be removed for ablation study, and we term the new variant of EDA as EDA-SW. The other components in the proposed EDA are necessary parts, and they do not have appropriate alternatives for an ablation study. 

The AOR and AR results obtained by the proposed EDA and its variant EDA-SW on the eight sequences in ECD and EED are shown in Fig. \ref{fig:ablation}. EDA-SW respectively achieves AOR/AR scores of 0.763/0.938, 0.757/0.953, 0.749/0.883, 0.901/1.000, 0.713/0.934, 0.735/0.917, 0.676/0.807, and 0.592/0.647 on the eight sequences (i.e., from \emph{shapes\_translation} to \emph{occlusions}). In contrast, EDA respectively obtains 0.896/0.998, 0.866/0.998, 0.872/0.983, 0.969/1.000, 0.913/1.000, 0.894/0.966, 0.752/0.917, and 0.702/0.833 on the eight sequences. From the results, we can see that the second stage of the TSW algorithm contributes significantly on improving the performance. By using the second stage of the TSW algorithm, the majority of pseudo-outliers are removed. This effectively reduces the ambiguity in the model selection process.

\subsection{Parameter Analysis}
\label{subsec:parameteranalysis}
In addition, we perform a parameter analysis on the challenging \emph{light\_ variations} sequence. Firstly, we study the influence of different inlier noise scale $\tau$. The proposed EDA respectively achieves AOR/AR scores of 0.889/0.963, 0.892/0.966, 0.894/0.966, 0.894/0.966, and 0.883/0.957 when $\tau$ is set to 0.0025, 0.005, 0.01, 0.02, and 0.05. From the results, we can see that the performance of EDA is not very sensitive to the value of $\tau$, and a small value of $\tau$ (around 0.01) can work well even in challenging scenes. Then, we replace $\tau$ with the IKOSE estimator \cite{wang2011simultaneously}. Since IKOSE needs the inlier ratio $k$, we study the influence of different $k$. EDA respectively obtains AOR/AR scores of 0.859/0.947, 0.883/0.966, 0.898/0.966, 0.871/0.947, and 0.847/0.941 when $k$ is set to 0.0025, 0.005, 0.01, 0.02, and 0.05. The comparison between the EDA with a fixed $\tau$ and the EDA with IKOSE are shown in the left image of Fig. \ref{fig:parameter}. As shown in the figure, the EDA with IKOSE only achieves a small improvement when $k$ is set to 0.01, and its AOR/AR scores drops considerably when $k$ is set to the other values compared with the EDA with a fixed $\tau$. Therefore, the proposed EDA adopts the fixed noise scale $\tau$ to avoid the trivial parameter tuning. Finally, we study the influence of different numbers of the time slices $N_s$ on the performance of EDA. EDA respectively achieves AOR/AR scores of 0.853/0.931, 0.897/0.966, 0.894/0.966, 0.769/0.941, and 0.691/0.875 when $N_s$ is set to 3, 5, 10, 15, and 20, as shown in the right image of Fig. \ref{fig:parameter}. The proposed EDA with $N_s$=5 only achieves 0.003 improvements on AOR compared with the EDA with $N_s$=10 at the cost of a 10x slower speed. Therefore, it is a good trade-off between accuracy and efficiency to set $N_s$ to 10.

\subsection{Time Cost}
\label{subsec:timecost}
The proposed EDA is evaluated on a PC with an Intel Core i7 CPU and an NVIDIA GTX 1080 GPU. For conventional tracking methods, they are synchronized with the global camera shutter, and thus their speeds are evaluated by a synchronous criterion (e.g., 25 frames per second and above can be considered as real-time). Since EDA works asynchronously, the synchronous criterion is not suitable for it. Instead, in event-based studies, the efficiency of event-based methods is commonly evaluated in terms of events per second (EPS) rather than the conventional frames per second (FPS). EDA runs at the average speed of 56.33K/31.72K EPS on the test sequences with/without the GPU support. Since the speed of event generation (about 30K EPS in relatively clutter-free scenes) is usually slower than the running speed of EDA (i.e., 56.33K/31.72K EPS), the proposed EDA is well suited for most common tasks.

\section{Conclusions}
In this paper, we propose a novel unifying event data association approach (EDA) to effectively and explicitly handle the essential event data association problem. The proposed EDA can asynchronously gather the event data over time for processing. Then, we present a deterministic model hypothesis generation strategy to effectively generate model hypotheses in the spatio-temporal domain from the gathered events. After that, we provide a two-stage weighting algorithm, which robustly weighs and selects the true models from the generated model hypotheses through multi-structural model fitting. Furthermore, we propose an adaptive model selection strategy to automatically determine the number of the true models. Finally, the selected true models are exploited to perform the final data association for remote object tracking. Extensive experiments on several challenging datasets demonstrate the effectiveness and superiority of the proposed EDA.

\vspace{6pt} 





\authorcontributions{Conceptualization, H.C. and Y.W.; methodology, H.C.; software, H.C. and Y.W.; validation, Y.W. and Y.P.; formal analysis, H.C.; investigation, H.C.; resources, H.C.; data curation, Y.W.; writing-original draft preparation, H.C.; writing—review and editing, Y.W. and Y.P.; visualization, Y.P.; supervision, Y.W. and Y.P. All authors have read and agreed to the published version of the manuscript.}

\funding{This work was supported by the China Postdoctoral Science Foundation (Grant Nos. GZC20233362, 2024MD754043, and 2024MD754039), Chongqing Postdoctoral Innovative Talents Support Program (Grant No. CQBX202316), Science and Technology Research Program of Chongqing Municipal Education Commission (Grant No. KJQN202400648), and National Natural Science Foundation of China (Grant No. 62301103).}

\dataavailability{The original contributions presented in this study are included in the article. Further inquiries can be directed to the corresponding author.}

\acknowledgments{The authors would like to thank all the reviewers for their valuable contributions to our work.}

\conflictsofinterest{The authors declare no conflict of interest.} 

\begin{adjustwidth}{-\extralength}{0cm}

\reftitle{References}


\bibliography{citations}

\begin{thebibliography}{999}

\bibitem[{\"O}lveczky et~al.(2003){\"O}lveczky, Baccus, and
  Meister]{olveczky2003segregation}
{\"O}lveczky, B.P.; Baccus, S.A.; Meister, M.
\newblock Segregation of object and background motion in the retina.
\newblock {\em Nature} {\bf 2003}, {\em 423},~401--408.

\bibitem[Lichtsteiner et~al.(2008)Lichtsteiner, Posch, and
  Delbruck]{lichtsteiner2008128}
Lichtsteiner, P.; Posch, C.; Delbruck, T.
\newblock A 128$\times$128 120 d{B} 15 $\mu$s latency asynchronous temporal
  contrast vision sensor.
\newblock {\em IEEE Journal of Solid-State Circuits} {\bf 2008}, {\em
  43},~566--576.

\bibitem[Brandli et~al.(2014)Brandli, Berner, Yang, Liu, and
  Delbruck]{brandli2014240}
Brandli, C.; Berner, R.; Yang, M.; Liu, S.C.; Delbruck, T.
\newblock A 240$\times$ 180 130 d{B} 3 $\mu$s latency global shutter
  spatiotemporal vision sensor.
\newblock {\em IEEE Journal of Solid-State Circuits} {\bf 2014}, {\em
  49},~2333--2341.

\bibitem[Posch et~al.(2011)Posch, Matolin, and Wohlgenannt]{posch2011qvga}
Posch, C.; Matolin, D.; Wohlgenannt, R.
\newblock A {QVGA} 143 d{B} dynamic range frame-free {PWM} image sensor with
  lossless pixel-level video compression and time-domain {CDS}.
\newblock {\em IEEE Journal of Solid-State Circuits} {\bf 2011}, {\em
  46},~259--275.

\bibitem[Gallego et~al.(2020)Gallego, Delbr{\"u}ck, Orchard, Bartolozzi, Taba,
  Censi, Leutenegger, Davison, Conradt, Daniilidis, et~al.]{gallego2020event}
Gallego, G.; Delbr{\"u}ck, T.; Orchard, G.; Bartolozzi, C.; Taba, B.; Censi,
  A.; Leutenegger, S.; Davison, A.J.; Conradt, J.; Daniilidis, K.;  et~al.
\newblock Event-based vision: A survey.
\newblock {\em IEEE Transactions on Pattern Analysis and Machine Intelligence}
  {\bf 2020}, {\em 44},~154--180.

\bibitem[Gallego et~al.(2018)Gallego, Rebecq, and
  Scaramuzza]{gallego2018unifying}
Gallego, G.; Rebecq, H.; Scaramuzza, D.
\newblock A unifying contrast maximization framework for event cameras, with
  applications to motion, depth, and optical flow estimation.
\newblock In Proceedings of the CVPR,  2018, pp. 3867--3876.

\bibitem[Zhu et~al.(2017)Zhu, Atanasov, and Daniilidis]{zhu2017event}
Zhu, A.Z.; Atanasov, N.; Daniilidis, K.
\newblock Event-based feature tracking with probabilistic data association.
\newblock In Proceedings of the ICRA,  2017, pp. 4465--4470.

\bibitem[Gallego et~al.(2019)Gallego, Gehrig, and Scaramuzza]{gallego2019focus}
Gallego, G.; Gehrig, M.; Scaramuzza, D.
\newblock Focus is all you need: Loss functions for event-based vision.
\newblock In Proceedings of the CVPR,  2019, pp. 12280--12289.

\bibitem[Xu et~al.(2024)Xu, Sun, and Wang]{xu2024sat}
Xu, W.; Sun, H.; Wang, S.
\newblock SAT: Spectrum-Adaptive Transformer with Spatial Awareness for UAV
  Target Tracking.
\newblock {\em Remote Sensing} {\bf 2024}, {\em 17},~52.

\bibitem[Yang et~al.(2024)Yang, Yang, and Feng]{yang2024siamrhic}
Yang, A.; Yang, Z.; Feng, W.
\newblock SiamRhic: Improved Cross-Correlation and Ranking Head-Based Siamese
  Network for Object Tracking in Remote Sensing Videos.
\newblock {\em Remote Sensing} {\bf 2024}, {\em 16},~4549.

\bibitem[Huang et~al.(2024)Huang, Huang, Niu, Miah, Wang, and
  Gao]{huang2024uav}
Huang, Y.; Huang, H.; Niu, M.; Miah, M.S.; Wang, H.; Gao, T.
\newblock UAV Complex-Scene Single-Target Tracking Based on Improved
  Re-Detection Staple Algorithm.
\newblock {\em Remote Sensing} {\bf 2024}, {\em 16},~1768.

\bibitem[Jiang and Yin(2023)]{jiang2023attention}
Jiang, Y.; Yin, G.
\newblock Attention-Enhanced One-Shot Attack against Single Object Tracking for
  Unmanned Aerial Vehicle Remote Sensing Images.
\newblock {\em Remote Sensing} {\bf 2023}, {\em 15},~4514.

\bibitem[Ye et~al.(2024)Ye, Xiao, and Liu]{ye2024multimodal}
Ye, P.; Xiao, G.; Liu, J.
\newblock Multimodal Features Alignment for Vision--Language Object Tracking.
\newblock {\em Remote Sensing} {\bf 2024}, {\em 16},~1168.

\bibitem[Liang et~al.(2024)Liang, Chen, Wu, Xia, and Li]{liang2024joint}
Liang, Y.; Chen, H.; Wu, Q.; Xia, C.; Li, J.
\newblock Joint spatio-temporal similarity and discrimination learning for
  visual tracking.
\newblock {\em IEEE Transactions on Circuits and Systems for Video Technology}
  {\bf 2024}.

\bibitem[Wang et~al.(2023)Wang, Jin, He, Huo, Wang, and Sun]{wang2023detector}
Wang, H.; Jin, L.; He, Y.; Huo, Z.; Wang, G.; Sun, X.
\newblock Detector--tracker integration framework for autonomous vehicles
  pedestrian tracking.
\newblock {\em Remote Sensing} {\bf 2023}, {\em 15},~2088.

\bibitem[Hou et~al.(2023)Hou, Cui, Ren, Li, Wang, and Jiao]{hou2023siamese}
Hou, B.; Cui, Y.; Ren, Z.; Li, Z.; Wang, S.; Jiao, L.
\newblock Siamese Multi-Scale Adaptive Search Network for Remote Sensing
  Single-Object Tracking.
\newblock {\em Remote Sensing} {\bf 2023}, {\em 15},~4359.

\bibitem[Liang et~al.(2021)Liang, Liu, Yan, Zhang, and Wang]{Liang2020robust}
Liang, Y.; Liu, Y.; Yan, Y.; Zhang, L.; Wang, H.
\newblock Robust visual tracking via spatio-temporal adaptive and channel
  selective correlation filters.
\newblock {\em Pattern Recognition} {\bf 2021}, {\em 112},~107738.

\bibitem[Gao et~al.(2025)Gao, Chen, Jiang, Xi, Xie, and Li]{gao2025feature}
Gao, L.; Chen, L.; Jiang, Y.; Xi, B.; Xie, W.; Li, Y.
\newblock Feature-level fusion network for hyperspectral object tracking via
  mixed multi-head self-attention learning.
\newblock {\em Remote Sensing} {\bf 2025}, {\em 17},~997.

\bibitem[Guo et~al.(2024)Guo, Li, An, Wang, He, Luo, Ling, Li, and
  Lin]{guo2024sptrack}
Guo, G.; Li, Z.; An, W.; Wang, Y.; He, X.; Luo, Y.; Ling, Q.; Li, M.; Lin, Z.
\newblock SPTrack: Spectral Similarity Prompt Learning for Hyperspectral Object
  Tracking.
\newblock {\em Remote Sensing} {\bf 2024}, {\em 16},~2975.

\bibitem[Xue et~al.(2023)Xue, Zhang, Lin, Li, Huo, and Zhang]{xue2023siamcaf}
Xue, Y.; Zhang, J.; Lin, Z.; Li, C.; Huo, B.; Zhang, Y.
\newblock SiamCAF: Complementary attention fusion-based Siamese network for
  RGBT tracking.
\newblock {\em Remote Sensing} {\bf 2023}, {\em 15},~3252.

\bibitem[Barranco et~al.(2018)Barranco, Fermuller, and Ros]{barranco2018real}
Barranco, F.; Fermuller, C.; Ros, E.
\newblock Real-time clustering and multi-target tracking using event-based
  sensors.
\newblock In Proceedings of the IROS,  2018, pp. 5764--5769.

\bibitem[Camu{\~n}as-Mesa et~al.(2017)Camu{\~n}as-Mesa, Serrano-Gotarredona,
  Ieng, Benosman, and Linares-Barranco]{camunas2017event}
Camu{\~n}as-Mesa, L.A.; Serrano-Gotarredona, T.; Ieng, S.H.; Benosman, R.;
  Linares-Barranco, B.
\newblock Event-driven stereo visual tracking algorithm to solve object
  occlusion.
\newblock {\em IEEE Transactions on Neural Networks and Learning Systems} {\bf
  2017}, {\em 29},~4223--4237.

\bibitem[Glover and Bartolozzi(2017)]{glover2017robust}
Glover, A.; Bartolozzi, C.
\newblock Robust visual tracking with a freely-moving event camera.
\newblock In Proceedings of the IROS,  2017, pp. 3769--3776.

\bibitem[Mitrokhin et~al.(2018)Mitrokhin, Ferm{\"u}ller, Parameshwara, and
  Aloimonos]{mitrokhin2018event}
Mitrokhin, A.; Ferm{\"u}ller, C.; Parameshwara, C.; Aloimonos, Y.
\newblock Event-based moving object detection and tracking.
\newblock In Proceedings of the IROS,  2018, pp. 1--9.

\bibitem[Chen et~al.(2019)Chen, Wu, Liang, Gao, and Wang]{chen2019asynchronous}
Chen, H.; Wu, Q.; Liang, Y.; Gao, X.; Wang, H.
\newblock Asynchronous tracking-by-detection on adaptive time surfaces for
  event-based object tracking.
\newblock In Proceedings of the ACM MM,  2019, pp. 473--481.

\bibitem[Chen et~al.(2020)Chen, Suter, Wu, and Wang]{chen2020end}
Chen, H.; Suter, D.; Wu, Q.; Wang, H.
\newblock End-to-end learning of object motion estimation from retinal events
  for event-based object tracking.
\newblock In Proceedings of the AAAI,  2020, pp. 10534--10541.

\bibitem[Zhu et~al.(2019)Zhu, Yuan, Chaney, and Daniilidis]{Zhu_2019_CVPR}
Zhu, A.Z.; Yuan, L.; Chaney, K.; Daniilidis, K.
\newblock Unsupervised event-based learning of optical flow, depth, and
  egomotion.
\newblock In Proceedings of the CVPR,  June 2019, pp. 989--997.

\bibitem[Gallego et~al.(2017)Gallego, Lund, Mueggler, Rebecq, Delbruck, and
  Scaramuzza]{gallego2017event}
Gallego, G.; Lund, J.E.; Mueggler, E.; Rebecq, H.; Delbruck, T.; Scaramuzza, D.
\newblock Event-based, 6-{DOF} camera tracking from photometric depth maps.
\newblock {\em IEEE Transactions on Pattern Analysis and Machine Intelligence}
  {\bf 2017}, {\em 40},~2402--2412.

\bibitem[Stoffregen et~al.(2019)Stoffregen, Gallego, Drummond, Kleeman, and
  Scaramuzza]{stoffregen2019event}
Stoffregen, T.; Gallego, G.; Drummond, T.; Kleeman, L.; Scaramuzza, D.
\newblock Event-based motion segmentation by motion compensation.
\newblock In Proceedings of the ICCV,  2019, pp. 7244--7253.

\bibitem[Clady et~al.(2015)Clady, Ieng, and Benosman]{clady2015asynchronous}
Clady, X.; Ieng, S.H.; Benosman, R.
\newblock Asynchronous event-based corner detection and matching.
\newblock {\em Neural Networks} {\bf 2015}, {\em 66},~91--106.

\bibitem[Lagorce et~al.(2017)Lagorce, Orchard, Galluppi, Shi, and
  Benosman]{lagorce2017hots}
Lagorce, X.; Orchard, G.; Galluppi, F.; Shi, B.E.; Benosman, R.B.
\newblock Hots: {A} hierarchy of event-based time-surfaces for pattern
  recognition.
\newblock {\em IEEE Transactions on Pattern Analysis and Machine Intelligence}
  {\bf 2017}, {\em 39},~1346--1359.

\bibitem[Fischler and Bolles(1981)]{fischler1981random}
Fischler, M.A.; Bolles, R.C.
\newblock Random sample consensus: a paradigm for model fitting with
  applications to image analysis and automated cartography.
\newblock {\em Communications of the ACM} {\bf 1981}, {\em 24},~381--395.

\bibitem[Pham et~al.(2014)Pham, Chin, Schindler, and
  Suter]{pham2014interacting}
Pham, T.T.; Chin, T.J.; Schindler, K.; Suter, D.
\newblock Interacting geometric priors for robust multimodel fitting.
\newblock {\em IEEE Transactions on Image Processing} {\bf 2014}, {\em
  23},~4601--4610.

\bibitem[Barath and Matas(2018)]{barath2018graph}
Barath, D.; Matas, J.
\newblock Graph-cut RANSAC.
\newblock In Proceedings of the CVPR,  2018, pp. 6733--6741.

\bibitem[Magri and Fusiello(2014)]{magri2014t}
Magri, L.; Fusiello, A.
\newblock T-linkage: A continuous relaxation of j-linkage for multi-model
  fitting.
\newblock In Proceedings of the CVPR,  2014, pp. 3954--3961.

\bibitem[Magri and Fusiello(2019)]{magri2019fitting}
Magri, L.; Fusiello, A.
\newblock Fitting multiple heterogeneous models by multi-class cascaded
  t-linkage.
\newblock In Proceedings of the CVPR,  2019, pp. 7460--7468.

\bibitem[Xiao et~al.(2020)Xiao, Ma, Wang, and Chen]{xiao2020deterministic}
Xiao, G.; Ma, J.; Wang, S.; Chen, C.
\newblock Deterministic model fitting by local-neighbor preservation and
  global-residual optimization.
\newblock {\em IEEE Transactions on Image Processing} {\bf 2020}, {\em
  29},~8988--9001.

\bibitem[Xiao et~al.(2016)Xiao, Wang, Lai, and Suter]{xiao2016hypergraph}
Xiao, G.; Wang, H.; Lai, T.; Suter, D.
\newblock Hypergraph modelling for geometric model fitting.
\newblock {\em Pattern Recognition} {\bf 2016}, {\em 60},~748--760.

\bibitem[Wang et~al.(2019)Wang, Xiao, Yan, and Suter]{wang2019searching}
Wang, H.; Xiao, G.; Yan, Y.; Suter, D.
\newblock Searching for representative modes on hypergraphs for robust
  geometric model fitting.
\newblock {\em IEEE Transactions on Pattern Analysis and Machine Intelligence}
  {\bf 2019}, {\em 41},~697--711.

\bibitem[Xiao et~al.(2019)Xiao, Wang, Yan, and Suter]{xiao2019superpixel}
Xiao, G.; Wang, H.; Yan, Y.; Suter, D.
\newblock Superpixel-guided two-view deterministic geometric model fitting.
\newblock {\em International Journal of Computer Vision} {\bf 2019}, {\em
  127},~323--339.

\bibitem[Ma et~al.(2019)Ma, Zhao, Jiang, Zhou, and Guo]{ma2019locality}
Ma, J.; Zhao, J.; Jiang, J.; Zhou, H.; Guo, X.
\newblock Locality preserving matching.
\newblock {\em International Journal of Computer Vision} {\bf 2019}, {\em
  127},~512--531.

\bibitem[Ma et~al.(2021)Ma, Jiang, Fan, Jiang, and Yan]{ma2020image}
Ma, J.; Jiang, X.; Fan, A.; Jiang, J.; Yan, J.
\newblock Image matching from handcrafted to deep features: A survey.
\newblock {\em International Journal of Computer Vision} {\bf 2021}, {\em
  129},~23--79.

\bibitem[Wang et~al.(2011)Wang, Chin, and Suter]{wang2011simultaneously}
Wang, H.; Chin, T.J.; Suter, D.
\newblock Simultaneously fitting and segmenting multiple-structure data with
  outliers.
\newblock {\em IEEE Transactions on Pattern Analysis and Machine Intelligence}
  {\bf 2011}, {\em 34},~1177--1192.

\bibitem[Mueggler et~al.(2017)Mueggler, Rebecq, Gallego, Delbruck, and
  Scaramuzza]{mueggler2017event}
Mueggler, E.; Rebecq, H.; Gallego, G.; Delbruck, T.; Scaramuzza, D.
\newblock The event-camera dataset and simulator: {E}vent-based data for pose
  estimation, visual odometry, and {SLAM}.
\newblock {\em International Journal of Robotics Research} {\bf 2017}, {\em
  36},~142--149.

\bibitem[Bertinetto et~al.(2016)Bertinetto, Valmadre, Henriques, Vedaldi, and
  Torr]{bertinetto2016fully}
Bertinetto, L.; Valmadre, J.; Henriques, J.F.; Vedaldi, A.; Torr, P.H.
\newblock Fully-convolutional siamese networks for object tracking.
\newblock In Proceedings of the ECCV,  2016, pp. 850--865.

\bibitem[Danelljan et~al.(2017)Danelljan, Bhat, Khan, Felsberg,
  et~al.]{danelljan2017eco}
Danelljan, M.; Bhat, G.; Khan, F.S.; Felsberg, M.;  et~al.
\newblock ECO: Efficient convolution operators for tracking.
\newblock In Proceedings of the CVPR,  2017, pp. 6638--6646.

\bibitem[Li et~al.(2019)Li, Wu, Wang, Zhang, Xing, and Yan]{Li_2019_CVPR}
Li, B.; Wu, W.; Wang, Q.; Zhang, F.; Xing, J.; Yan, J.
\newblock SiamRPN++: Evolution of siamese visual tracking with very deep
  networks.
\newblock In Proceedings of the CVPR,  2019, pp. 4282--4291.

\bibitem[Danelljan et~al.(2019)Danelljan, Bhat, Khan, and
  Felsberg]{Danelljan_2019_CVPR}
Danelljan, M.; Bhat, G.; Khan, F.S.; Felsberg, M.
\newblock ATOM: Accurate tracking by overlap maximization.
\newblock In Proceedings of the CVPR,  2019, pp. 4660--4669.

\bibitem[Scheerlinck et~al.(2019)Scheerlinck, Rebecq, Stoffregen, Barnes,
  Mahony, and Scaramuzza]{scheerlinck2019ced}
Scheerlinck, C.; Rebecq, H.; Stoffregen, T.; Barnes, N.; Mahony, R.;
  Scaramuzza, D.
\newblock CED: Color event camera dataset.
\newblock In Proceedings of the CVPRW,  2019, pp. 1684--1693.

\bibitem[Mitrokhin et~al.(2019)Mitrokhin, Ye, Fermuller, Aloimonos, and
  Delbruck]{mitrokhin2019ev}
Mitrokhin, A.; Ye, C.; Fermuller, C.; Aloimonos, Y.; Delbruck, T.
\newblock EV-IMO: Motion segmentation dataset and learning pipeline for event
  cameras.
\newblock In Proceedings of the IROS,  2019, pp. 6105--6112.

\end{thebibliography}

%


\PublishersNote{}
\end{adjustwidth}
\end{document}